\documentclass{article}
\usepackage{spconf,amsmath,graphicx}
\usepackage{booktabs} 
\usepackage{bbding}
\usepackage{amssymb}
\usepackage{xcolor}
\usepackage{pifont}
\usepackage{microtype}

\usepackage[pagebackref=false,breaklinks=true,letterpaper=true,colorlinks,bookmarks=false,citecolor=citecolor,linkcolor=linkcolor]{hyperref}

\definecolor{citecolor}{HTML}{0071BC}
\definecolor{linkcolor}{HTML}{ED1C24}

\newcommand{\cmark}{\ding{51}}

\definecolor{brickred}{rgb}{0.8, 0.25, 0.33}
\definecolor{blueish}{rgb}{0.0, 0.3, 0.6}

\title{\hspace{-1.6mm}Interweaved\hspace{-0.1mm} Graph\hspace{-0.1mm} and\hspace{-0.1mm} Attention\hspace{-0.1mm} Network\hspace{-0.1mm} for\hspace{-0.1mm} 3D\hspace{-0.1mm} Human\hspace{-0.1mm} Pose\hspace{-0.1mm} Estimation\hspace{-1.6mm}}
\name{Ti Wang$^1$, Hong Liu$^1$, Runwei Ding$^{1*}$, Wenhao Li$^1$, Yingxuan You$^1$, Xia Li$^2$ \thanks{*Corresponding author: dingrunwei@pku.edu.cn. This work is supported by National Natural Science Foundation of China (No.62073004), Basic and Applied Basic Research Foundation of Guangdong (No.2020A1515110370), Shenzhen Fundamental Research Program (GXWD20201231165807007-20200807164903001, No.JCYJ20200109140410340).}
}

\address{
Key Laboratory of Machine Perception, Peking University, Shenzhen Graduate School$^1$ \\
Department of Computer Science, ETH Zürich$^2$\\
\texttt{\small\{tiwang, youyx\}@stu.pku.edu.cn, \{hongliu, dingrunwei, wenhaoli\}@pku.edu.cn, xia.li@inf.ethz.ch}
}

\begin{document}

\maketitle
\begin{abstract}
Despite substantial progress in 3D human pose estimation from a single-view image, prior works rarely explore global and local correlations, leading to insufficient learning of human skeleton representations. To address this issue, we propose a novel Interweaved Graph and Attention Network (IGANet) that allows bidirectional communications between graph convolutional networks (GCNs) and attentions. Specifically, we introduce an IGA module, where attentions are provided with local information from GCNs and GCNs are injected with global information from attentions. Additionally, we design a simple yet effective U-shaped multi-layer perceptron (uMLP), which can capture multi-granularity information for body joints. Extensive experiments on two popular benchmark datasets (i.e. Human3.6M and MPI-INF-3DHP) are conducted to evaluate our proposed method.The results show that IGANet achieves state-of-the-art performance on both datasets. Code is available at \href{https://github.com/xiu-cs/IGANet}{https://github.com/xiu-cs/IGANet}.
\end{abstract}

\begin{keywords}
3D Human Pose Estimation, Graph Convolutional Network, Attention
\end{keywords}

\vspace{-1mm}
\section{Introduction}\label{sec:intro}
\vspace{-1mm}
Monocular 3D human pose estimation aims to recover the 3D positions of body joints from a single-view image.
This task plays an important role in many applications, such as human-computer interaction, action recognition, and human mesh reconstruction. 
Typically, the pipeline can be divided into two parts: 1) estimating the locations of 2D keypoints from a monocular image, and 2) lifting the estimated 2D keypoints to 3D.
In this paper, we focus on the problem of 2D-3D pose lifting, where the model input is a 2D pose detected from an image using off-the-shelf 2D pose detectors~\cite{chen2018cascaded,newell2016stacked}.

With the development of graph convolutional networks (GCNs) in recent years, various GCN-based methods~\cite{zhao2019semantic, zou2021modulated} combined with human geometry priors have significantly improved the accuracy of predicting 3D human skeleton positions.
For a given graph node, GCN-based methods can exploit semantic information by aggregating the features of its neighbor nodes.
However, some distant joints can also provide useful information, such as the relations between the joints of hands and feet when running, which are difficult to capture by the GCN-based methods.
Recently, transformers have been widely applied to 3D human pose estimation~\cite{li2022exploiting,zheng20213d,li2022mhformer}. 
Thanks to the attention mechanism of transformers~\cite{vaswani2017attention}, global relations between body joints can be well captured.
However, without the topological structure information of the human skeleton, attentions can not fully explore the geometrical connections between body joints. 
There are few existing works that combine GCNs and attentions in a serial or parallel manner~\cite{lin2021mesh,zheng2022lightweight,li2022graphmlp}. 
Nevertheless, these simple combination manners cannot sufficiently utilize the complementarity of GCNs and attentions for effective global and local modeling.

\begin{figure}[!t]\label{fig:overview_model}
    \centering
    \includegraphics[width=0.7 \linewidth]{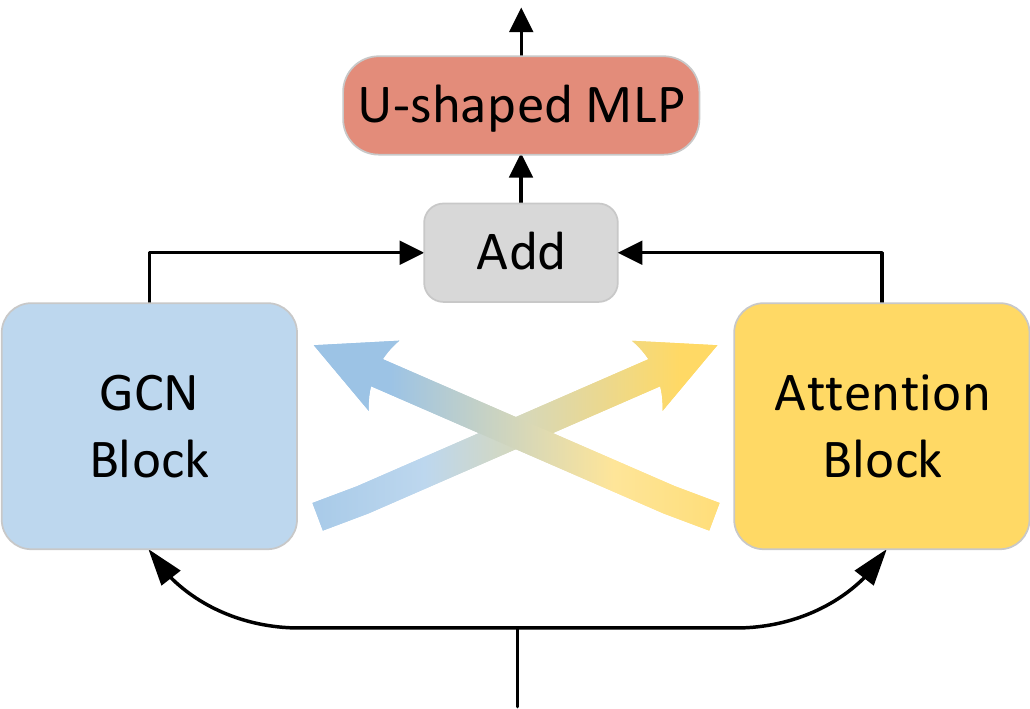}
    \caption{A schematic diagram of IGANet. The local relations extracted by the GCN block are interweaved with the global information obtained by the attention block.}
    \label{fig:moti}
    \vspace{-6mm}
\end{figure}

\begin{figure*}[!t]
    \centering
    \centerline{\includegraphics[width=1\linewidth]{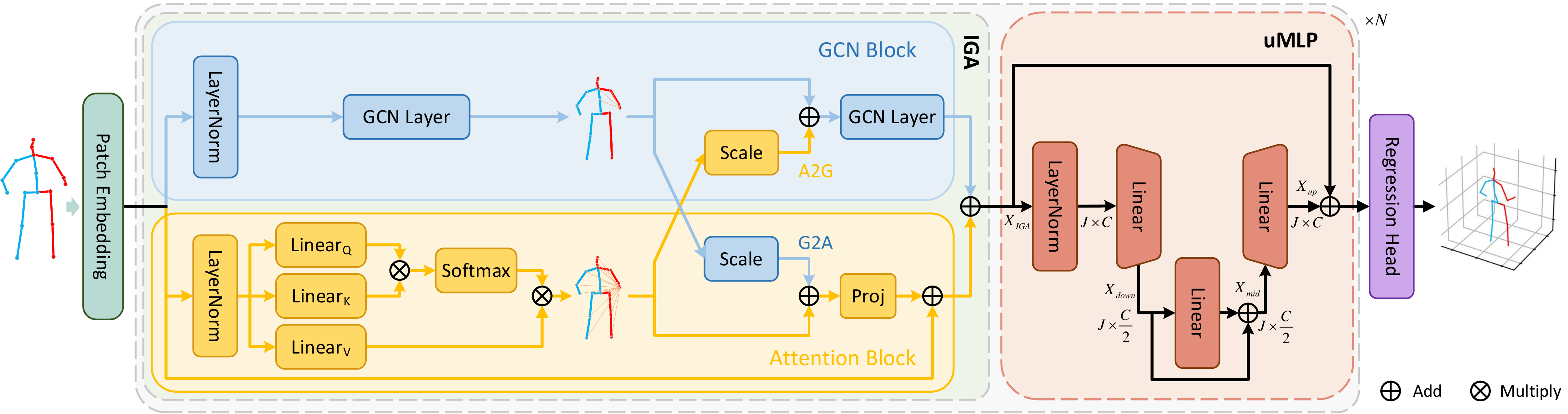}}
    \caption
    {
        Structures of Interweaved Graph and Attention Network (IGANet). 
        The input 2D body joints are embedded into high-dimensional features through the patch embedding.
        In our IGA module, the captured global and local human skeleton features are interweaved to provide complementary clues.
        After that, our uMLP module further captures multi-granularity information of body joints.
        Finally, the predicted results are obtained by the regression head.
    }
    \label{fig:methods}
    \vspace{-1mm}
\end{figure*}

To overcome this issue, a novel Interweaved Graph and Attention Network (IGANet) is proposed to learn bidirectional communications between GCNs and attentions, as shown in Fig.~\ref{fig:overview_model}. 
Specifically, we introduce an IGA module based on two guidance strategies:
a) \textit{G2A}: the topology information of the human skeleton extracted by the GCN block is injected into the attention block, so that the attention can be guided by the GCN to better learn the structure information of the human body.
b) \textit{A2G}: the global information of body joints captured by the attention block is passed to the GCN block, so that the GCN can take the global connections into account while paying attention to its neighbor nodes. 
Moreover, inspired by the U-shaped structure~\cite{ronneberger2015u,cai2019exploiting} that captures features at multiple scale, we design a U-shaped multi-layer perception (uMLP) module with a bottleneck structure along the channel dimension, which can learn multi-granularity features via down-projection and up-projection layers. 
We find that such a simple design outperforms the original MLP in transformer for learning skeletal representations.

Overall, our contributions can be summarized as follows:
(1) We propose a new network named IGANet for single-frame 3D human pose estimation, which allows GCNs and attentions to complement each other. 
(2) A simple yet effective U-shaped MLP is designed to capture multi-granularity information.
(3) The proposed IGANet outperforms existing state-of-the-art methods on two popular benchmark datasets, \textit{i.e.}, Human3.6M~\cite{ionescu2013human3} and MPI-INF-3DHP~\cite{mehta2017monocular}.

\vspace{-1mm}
\section{Method}
\vspace{-1mm}

As shown in Fig.~\ref{fig:methods}, IGANet contains two main components: the Interweaved Graph and Attention module (IGA) and the U-shaped MLP module (uMLP). 
Through patch embedding, the 2D skeleton is mapped into $X {\in} \mathbb{R}^{J \times C}$, 
and then sent to IGA and uMLP to extract the features of body joints. 
Finally, regressions are performed to predict the 3D pose $Y {\in} \mathbb{R}^{J \times 3}$.
IGANet can learn bidirectional communications between GCNs and attentions to mutually improve each other, thus enhancing the capability of modeling human skeleton relationships.
% The main insight is to use the complementary advantages of GCN and attention to better learn the human skeleton relationship.
% The insight for using interweaved GCNs and attentions is to interchange complementary information with each other, thus enhancing the capability of modeling human skeleton representations.

\begin{table*}[t]
    % \footnotesize
    \caption{
        Quantitative comparisons on Human3.6M under MPJPE. The 2D pose detected by cascaded pyramid network (CPN)~\cite{chen2018cascaded} is used as input.
        We use $\S$ to highlight methods that use refinement module~\cite{zou2021modulated, cai2019exploiting}.
        The top two best results for each action are highlighted in bold and underlined, respectively.
    }
\footnotesize
\setlength\tabcolsep{1.65mm}
\begin{tabular}{l|ccccccccccccccc|c}
\toprule
Method & Dire. & Disc. & Eat & Greet & Phone & Photo & Pose & Purch. & Sit & SitD. & Smoke & Wait & WalkD. & Walk & WalkT. & Avg   \\
\midrule

Martinez \textit{et al.}~\cite{martinez2017simple} &51.8& 56.2& 58.1& 59.0&69.5&78.4& 55.2& 58.1& 74.0& 94.6&62.3&59.1& 65.1&49.5& 52.4& 62.9\\

Zhao \textit{et al.}~\cite{zhao2019semantic} & 47.3 & 60.7 & 51.4 & 60.5 & 61.1 & \textbf{49.9} & 47.3 & 68.1 & 86.2 & \textbf{55.0} & 67.8 & 61.0 & \textbf{42.1} & 60.6 & 45.3 & 57.6 \\
%  LCN
Ci \textit{et al.}~\cite{ci2019optimizing} & 46.8 & 52.3 & \underline{44.7} & 50.4 & 52.9 & 68.9 & 49.6 & 46.4 & 60.2 & 78.9 & 51.2 & 50.0 & 54.8 & 40.4 & 43.3 & 52.7\\
% GraphSH
Xu \textit{et al.}~\cite{xu2021graph} &  45.2&49.9 &47.5& 50.9 &54.9& 66.1& 48.5 &46.3 &59.7 &71.5& 51.4 &48.6 &53.9 &39.9 &44.1 &51.9 \\
Zhao \textit{et al.}~\cite{zhao2022graformer} & 45.2 &50.8 & 48.0 & 50.0 & 54.9& 65.0 & 48.2& 47.1&60.2&70.0&51.6&48.7&54.1&39.7&43.1&51.8 \\
% STGCN where * indicates single frame methods
Cai \textit{et al.}~\cite{cai2019exploiting}$\S$ & 46.5 & 48.8 & 47.6 & 50.9 & 52.9 & 61.3 & 48.3 & 45.8 & 59.2 & 64.4 & 51.2 & 48.4 & 53.5 & 39.2 & 41.2 & 50.6\\
% SRNet
Zeng \textit{et al.}~\cite{zeng2020srnet}& 44.5 & 48.2 & 47.1 & 47.8 & 51.2 & 56.8 & 50.1 & 45.6 & 59.9 & 66.4 & 52.1 & \underline{45.3} & 54.2 & 39.1 & 40.3 & 49.9\\

Zou \textit{et al.}~\cite{zou2021modulated}$\S$ & 45.4 & 49.2 & 45.7 & 49.4 & 50.4 & 58.2 & 47.9 & 46.0 & 57.5 & 63.0 & 49.7 & 46.6 & 52.2 & 38.9 & 40.8 & 49.4 \\
% Gong \textit{et al.}~\cite{gong2021poseaug} & -&-& -&-&- & -&  -&-&- & -& -& - &- &- &- & 50.2\\
\midrule
IGANet (Ours) & \underline{42.9}  & \underline{47.9} & \textbf{44.9} & \underline{47.7} & \underline{49.8} & 58.4 & \underline{46.0}  & \underline{44.7} & \underline{56.8} & 61.4 & \underline{49.3} & 46.1 & 52.0 & \underline{37.6} & \underline{39.8} & \underline{48.3}\\
IGANet (Ours)$\S$ & \textbf{42.6}  & \textbf{47.3} & 45.4 & \textbf{47.6} & \textbf{49.5} & \underline{56.2} & \textbf{45.9}  & \textbf{44.1} & \textbf{56.3} &  \underline{59.2} & \textbf{48.6} & \textbf{45.1} & \underline{50.3} & \textbf{37.3} & \textbf{39.6} & \textbf{47.7}\\

\bottomrule
\end{tabular}
\label{tab:human_cpn_p1}
\vspace{-2mm}
\end{table*}

\vspace{-1mm}
\subsection{Preliminary}
\label{sec:gcn}
\noindent \textbf{Graph Convolutional Network (GCN).}
% \noindent \textbf{GCN.}
An undirected graph can be represented as $G {=} \{V, E\}$, where $V$ is the set of nodes, and $E$ is the set of edges. 
The edges can be encoded in an adjacency matrix $A {\in} \{0,1\}^{N {\times} N}$. 
For the $l^{th}$ layer feature $X_{l}$, the vanilla graph convolution aggregates neighboring node features, which can be formulated as:
\begin{equation}
    X_{l} = \sigma(W_{l}X_{l-1}\tilde{A} ),
    \label{equ:E_gcn}
\end{equation}
where $W_{l} {\in} \mathbb{R}^{C^{'} \times C} $ is the layer-specific trainable weight matrix,
$\tilde{A} {=} A {+} I_N$ is the adjacency matrix of the graph with added self-connections, and $I_N$ is the identity matrix.
The stacking of multiple graph convolutional layers aggregates neighboring nodes to obtain enhanced feature representations.
% By stacking multiple graph convolutional layers, 
% GCN iteratively transforms and aggregates neighboring nodes to obtain enhanced feature representations.
% By stacking multiple graph convolutional layers, neighboring nodes are aggregated to obtain enhanced feature representations.

\noindent \textbf{Attention.}
\label{sec:attn}
The input tokens $X {\in} \mathbb{R}^{J \times C}$ are first projected to queries $Q {\in} \mathbb{R}^{J\times d}$, keys $K {\in} \mathbb{R}^{J\times d}$, and values $V {\in} \mathbb{R}^{J\times d}$,
and then $Q, K,V$ are fed to a scaled dot-product attention~\cite{vaswani2017attention}:
\begin{equation}
    \text{Attention}(Q,K,V)=\text{Softmax}(QK^{T}/\sqrt{d})V,
    \label{equ:MSA}
\end{equation}
where $d$ is the dimension of $Q, K, V$. Multi-head self-attention~(MSA)~\cite{vaswani2017attention} splits $Q$, $K$, $V$ into multiple heads, each of which applies scaled dot-product attention in parallel.
% The formula can be expressed as:
% \begin{equation}
%     \text{MSA}(Q,K,V)=\text{Concat}(H_1,H_2,...,H_h)W,
%     \label{equ:MSA}
% \end{equation}
% where $H_i {=} \text{Attention}(Q_i,K_i,V_i)$ denotes the $i^{th}{\in} [1,...,h]$ head, and $W$ is the projection transformation matrix.
This enables the model to efficiently utilizes information from various representation subspaces with different locations.

% \noindent \textbf{Attention.}
% \label{sec:attn}
% The input tokens $X {\in} \mathbb{R}^{J \times C}$ are first projected to queries $Q {\in} \mathbb{R}^{J\times d}$, keys $K {\in} \mathbb{R}^{J\times d}$, and values $V {\in} \mathbb{R}^{J\times d}$,
% and then $Q, K,V$ are fed to a scaled dot-product attention~\cite{vaswani2017attention}:
% \begin{equation}
%     \text{Attention}(Q,K,V)=\text{Softmax}(QK^{T}/\sqrt{d})V,
%     \label{equ:MSA}
% \end{equation}
% where $d$ is the dimension of $Q, K, V$.

\vspace{-1mm}
\subsection{Interweaved Graph and Attention}
% "The xxx module" or "xxx"
\label{sec:GIA}
% Our IGANet contains two major components: IGA module and U-shaped MLP. 
% As illustrated in Fig.~\ref{fig:methods}, the 2D skeletons after patch embedding are sent to GCN blocks and attention blocks respectively.
% Fig.~\ref{fig:methods} shows the architecture of IGANet. 
% After patch embedding, the 2D skeleton is sent to GCN block and attention block respectively.
The attention block can capture long-range relationships between nodes in the graph, 
but it is difficult to model the inherent patterns of the human body structure, such as the left-right symmetry of the human body. 
The GCN focuses on the local information of human body joints but is limited in capturing global dependencies. 
To relieve these limitations, we propose an IGA module, which contains a GCN block and an attention block with two guidance strategies~(\textit{i.e.}, Graph2Attention~(G2A) and Attention2Graph~(A2G)) to interchange complementary information with each other. 

\noindent \textbf{Graph2Attention (G2A).}  
To guide the attention to learn the topology priors of the human skeleton, we inject the skeletal information $f_{graph}$ captured by the first layer of the GCN block into the attention block.
This can be formulated as:
\begin{equation}
    X_{G2A} = \text{Softmax}(QK^{T}/\sqrt{d})V+s_{G2A}\cdot f_{graph},
    \label{equ:graph2attn}
\end{equation}
where $s_{G2A}$ is the scale factor for $f_{graph}$.
With the guidance of skeleton information from the GCN block,
attention can fully capture the relations between human body joints.

\noindent \textbf{Attention2Graph (A2G).}
Similarly, we feed back the global relations of human skeleton $f_{global}$ captured by the attention block into the GCN block,
which enables the GCN to have a better awareness of the global relations of human body joints.
This operation can be defined as:
\begin{equation}
    X_{A2G}= G_1 + s_{A2G}\cdot f_{global},
    \label{equ:G2A}
\end{equation}
% where $G_1$ indicates the output of the first layer of the GCN block,
where $G_1$ is the output of the first GCN layer in GCN block,
$s_{A2G}$ denotes the scale factor for $f_{global}$.
In this way, the global information can be perceived by the GCN block. 

With the interweaved complementary information, the modeling abilities of the GCN block and the attention block are both enhanced.
Finally, the outputs from these two blocks are added together, which can be expressed as:
\begin{equation}
    X_{IGA}=G_2(X_{A2G})+\text{Proj}(X_{G2A}),
    \label{equ:mix}
\end{equation}
where $\text{Proj}(\cdot)$ is the projection head containing a linear layer.

\begin{table}[!t]
    \centering
    \caption{Quantitative comparisons on Human3.6M under MPJPE. Ground truth 2D pose is used as input.}
    \setlength\tabcolsep{12.0mm}
    \footnotesize
    \begin{tabular}{l|c}
    \toprule

    Method & MPJPE $\downarrow$ \\
    \midrule
    Martinez \textit{et al.}~\cite{martinez2017simple} & 45.5 \\
    Zhao \textit{et al.}~\cite{zhao2019semantic} & 43.8\\
    Cai \textit{et al.}~\cite{cai2019exploiting} & 38.1\\
    Gong \textit{et al.}~\cite{gong2021poseaug}  & 38.2\\
    Zou \textit{et al.}~\cite{zou2021modulated} &  37.4\\
    Zeng \textit{et al.}~\cite{zeng2020srnet} & 36.4\\
   Xu \textit{et al.}~\cite{xu2021graph} & 35.8\\
    \midrule
    IGANet (Ours) & \textbf{32.7 ($\uparrow$ \scriptsize 8.7\%)} \\
    % \hline
    \bottomrule
    \end{tabular}
    \label{tab:Human_gt}
    \vspace{-1mm}
\end{table}

\begin{table}[!t]
    \centering
    \caption{Quantitative comparisons with state-of-the-art methods on MPI-INF-3DHP. }
    \setlength\tabcolsep{3.20mm}
    \footnotesize
    \begin{tabular}{l|c|c|c}
    \toprule 
    Method  & Outdoor & All (PCK) $\uparrow$& All (AUC) $\uparrow$\\
    \midrule
    Martinez \textit{et al.}~\cite{martinez2017simple} & 31.2&42.5& 17.0 \\
    Ci \textit{et al.}~\cite{ci2019optimizing} & 77.3 & 74.0 & 36.7 \\
    Li \textit{et al.}~\cite{li2019generating} & 66.6& 67.9& - \\
    Zeng \textit{et al.}~\cite{zeng2020srnet} & 80.3 & 77.6 & 43.8 \\
    Xu \textit{et al.}~\cite{xu2021graph} & 75.2 & 80.1 & 45.8 \\ 
    % Zeng \textit{et al.}~\cite{zeng2021learning} & - &- &84.6 & 82.1 & 46.2 \\
    Liu \textit{et al.}~\cite{liu2020comprehensive} & 80.1& 79.3& 47.6 \\
    Zou \textit{et al.}~\cite{zou2021modulated} & 85.7 & 86.1 & 53.7  \\
    \midrule
    % Our&  84.1 & 86.5 & 87.7 & 85.8 & 53.4 \\
    % OurV1.9.2 &  85.1 & 86.6 & 87.5 & 86.2 & 52.7  \\
    IGANet (Ours) & \textbf{86.4} & \textbf{86.1} & \textbf{54.2}  \\
    \bottomrule
    \end{tabular}
    \label{tab:3dhp}
    \vspace{-1mm}
\end{table}

\vspace{-1mm}
\subsection{U-shaped MLP}
% "The xxx module" or "xxx"
\label{sec:umlp}
Unlike the original MLP of transformer that uses an inverted bottleneck structure, we design a simple yet effective U-shaped MLP (uMLP) that adopts a bottleneck structure along the channel dimension.  
The input $X_{IGA}$ is first fed to a down-projection layer,
followed by a middle layer that keeps the same dimension, and finally to an up-projection layer.
To preserve the original information, shortcuts are used between layers of the same dimension.
This above is formulated as:
\begin{equation}
\begin{aligned}
    X_{down} &= \text{MLP}_{down}(\text{LN}(X_{IGA})), \\
    X_{mid} &= \text{MLP}_{mid}(X_{down})+X_{down} , \\
    X_{up} &= \text{MLP}_{up}(X_{mid})+X_{IGA},
    \label{equ:umlp}
\end{aligned}            
\end{equation}
where $\text{MLP} (\cdot)$ consists of a linear layer and a $\text{GELU}$ activation function, $\text{LN}(\cdot)$ denotes the Layer Normalization.

\vspace{-1mm}
\section{Experiments}
\vspace{-1mm}

\subsection{Datasets and Evaluation Metrics}
% \noindent \textbf{Human3.6M and MPI-INF-3DHP.}
% Human3.6M~\cite{ionescu2013human3} is the most widely used indoor dataset for 3D human pose estimation.
% Following previous works~\cite{pavllo20193d}, we train the model on 5 subjects (S1, S5, S6, S7, S8) and test on 2 subjects (S9 and S11). 
% To demonstrate the generalization ability of our model, we directly test the model trained through Human3.6M on a more challenging 3D pose dataset, i.e, MPI-INF-3DHP~\cite{mehta2017monocular}.
\noindent \textbf{Human3.6M.}
Human3.6M~\cite{ionescu2013human3} is the most widely used indoor dataset for 3D human pose estimation.
Following previous works~\cite{zou2021modulated,martinez2017simple,xu2021graph}, our model is trained on 5 subjects (S1, S5, S6, S7, S8) and tested on 2 subjects (S9 and S11). 
We adopt the most commonly used evaluation metric MPJPE~\cite{dang2019deep}, which calculates the mean Euclidean distance between the estimated joints and the ground truth in millimeters.

%  缩减版 MPI-INF-3DHP
\noindent \textbf{MPI-INF-3DHP.}
% To demonstrate the generalization ability of our model,
% we directly test the model trained with Human3.6M on the MPI-INF-3DHP~\cite{mehta2017monocular} datset that contains both indoor and complex outdoor scenes.
MPI-INF-3DHP~\cite{mehta2017monocular} is a more challenging dataset that contains both indoor and complex outdoor scenes.
This dataset records 8 actors performing 8 activity sets each from 14 camera views, covering more diverse poses than Human3.6M.
Following previous works~\cite{zou2021modulated,zeng2020srnet}, 
the percentage of correct keypoints (PCK) under 150 mm radius and the area under the curve (AUC) are used as evaluation metrics.

\subsection{Implementation Details}
% Our model is trained and tested on a single NVIDIA RTX 1080 Ti GPU.
% As depicted in Fig.~\ref{fig:methods}, 
% During the inference, the embeded inputs $X {\in} \mathbb{R}^{J \times C}$ with channel dimension $C {=} 512$ are sent to the block consisting of IGA and uMLP modules, looping $N {=} 3$ times.
% During the inference, the embeded inputs with channel dimension $C {=} 512$ are sent to IGA and uMLP modules, looping $N {=} 3$ times.
% The IGANet consisting of IGA and uMLP modules loops $N {=} 3$ times with channel dimension $C {=} 512$. 
Our IGANet containing IGA and uMLP modules loops $N {=} 3$ times.
The scale factor of $f_{global}$ and $f_{graph}$ are set as 0.8 and 0.5.
Following~\cite{zou2021modulated}, 
we adopt 2D joints detected by CPN~\cite{chen2018cascaded} on Human3.6M and ground truth 2D joints on MPI-INF-3DHP.
Horizontal flipping is used for data augmentation during training and testing following~\cite{zou2021modulated, cai2019exploiting}.
% We train our model for 20 epochs on a single RTX 1080 Ti GPU and set the batch size to 128.
% The Adam optimizer is adopted with an initial learning rate of 0.001 and decay factor of 0.95 per epoch.
Our model is trained for 20 epochs on a single RTX 1080 Ti GPU and the batch size is set to 128.
The Adam optimizer is adopted with an initial learning rate of 0.001 and a decay factor of 0.95 per epoch.

\vspace{-1mm}
\subsection{Comparison with State-of-the-Art Methods}
\noindent \textbf{Human3.6M.}
Table~\ref{tab:human_cpn_p1} shows the comparisons of our IGANet with previous single-frame methods, using 2D keypoints detected by CPN~\cite{chen2018cascaded}. 
IGANet significantly improves the performance from 49.9 mm to 48.3 mm in MPJPE.
We note that some works~\cite{zou2021modulated, cai2019exploiting} adopt refinement module to further boost the performance. 
Following them, IGANet achieves 47.7 mm MPJPE surpassing MGCN~\cite{zou2021modulated} by a large margin (relative 3.4\% improvement).
% Our method achieves optimal or sub-optimal performance on all actions.
% It can be seen that our method achieves optimal or sub-optimal performance on all actions.
% Due to the uncertainty in the detected 2D poses, 
% We further use the ground truth keypoints as input to explore the upper bound of the proposed method.
% The results are shown in Table~\ref{tab:Human_gt}.
%  an obvious ->a noticeable
% Compared with previous methods, our method can achieve an obvious improvement (from 35.8mm to 32.7mm, relative 8.7\% improvements),
% which also proves the effectiveness of our method.
% Comparision results in Table~\ref{tab:Human_gt} show that our method can achieve an obvious improvement (from 35.8mm to 32.7mm, relative 8.7\% improvements),which .
% We further use the ground truth keypoints as input 
To explore the upper bound of our method, comparison results using ground truth 2D keypoints are shown in Table~\ref{tab:Human_gt}.
Our method achieves an obvious improvement (from 35.8 mm to 32.7 mm, relative 8.7\% improvements), which further demonstrate its effectiveness.

\begin{table}  
    \footnotesize
    \centering
    \caption{
    Ablation study on different designs of our model.
    }
    \setlength\tabcolsep{3.05mm}
    \footnotesize
    \begin{tabular}{c|c|cc|c|c}    
    \toprule
     Attention& GCN  &  G2A  &   A2G  &  uMLP & MPJPE $\downarrow$ \\
    \midrule
     \cmark &     &  &  &   &51.7  \\ % attn +MLP
    % \hline
    % & \cmark & & & & \cmark &52.6  \\ % attn +uMLP
    %\CheckmarkBold &    &  & \CheckmarkBold &      &   \\ % gcn+mlp
    % \cmark &  &  & &  &  \cmark & 49.4  \\ % gcn + umlp    
    \cmark & \cmark &   &  &   & 49.6  \\ % gcn + attn + MLP    
    % \cmark & \cmark &   &  &  \cmark & 49.3  \\ % gcn + attn + umlp
    \cmark & \cmark & \cmark &  &  & 49.2  \\ % gcn+attn+G2A+mlP
    \cmark & \cmark &  & \cmark &  & 49.4  \\ % gcn+attn+A2G+mlp  
    % \cmark & \cmark  & \cmark &\cmark &   &   49.1\\ % gcn+attn+GIA+mlp
    \cmark & \cmark & \cmark & & \cmark  & 48.8  \\ % gcn+attn+G2A+uMLP
    \cmark & \cmark &  & \cmark & \cmark  & 48.8  \\ % gcn+attn+A2G+uMLP
    %   \hline
    % \cmark & \cmark  & \cmark &\cmark &   &   49.1\\ % gcn+attn+GIA+mlp
    \midrule
    \cmark & \cmark & \cmark & \cmark & \cmark  & \textbf{48.3}  \\ %gcn+attn+GIA+umlp
    \bottomrule
    \end{tabular}    
    \label{tab:ab_study}
    \vspace{-1mm}
\end{table}

\begin{figure}[!t]
    \centering
    \includegraphics[width=1.0\linewidth]{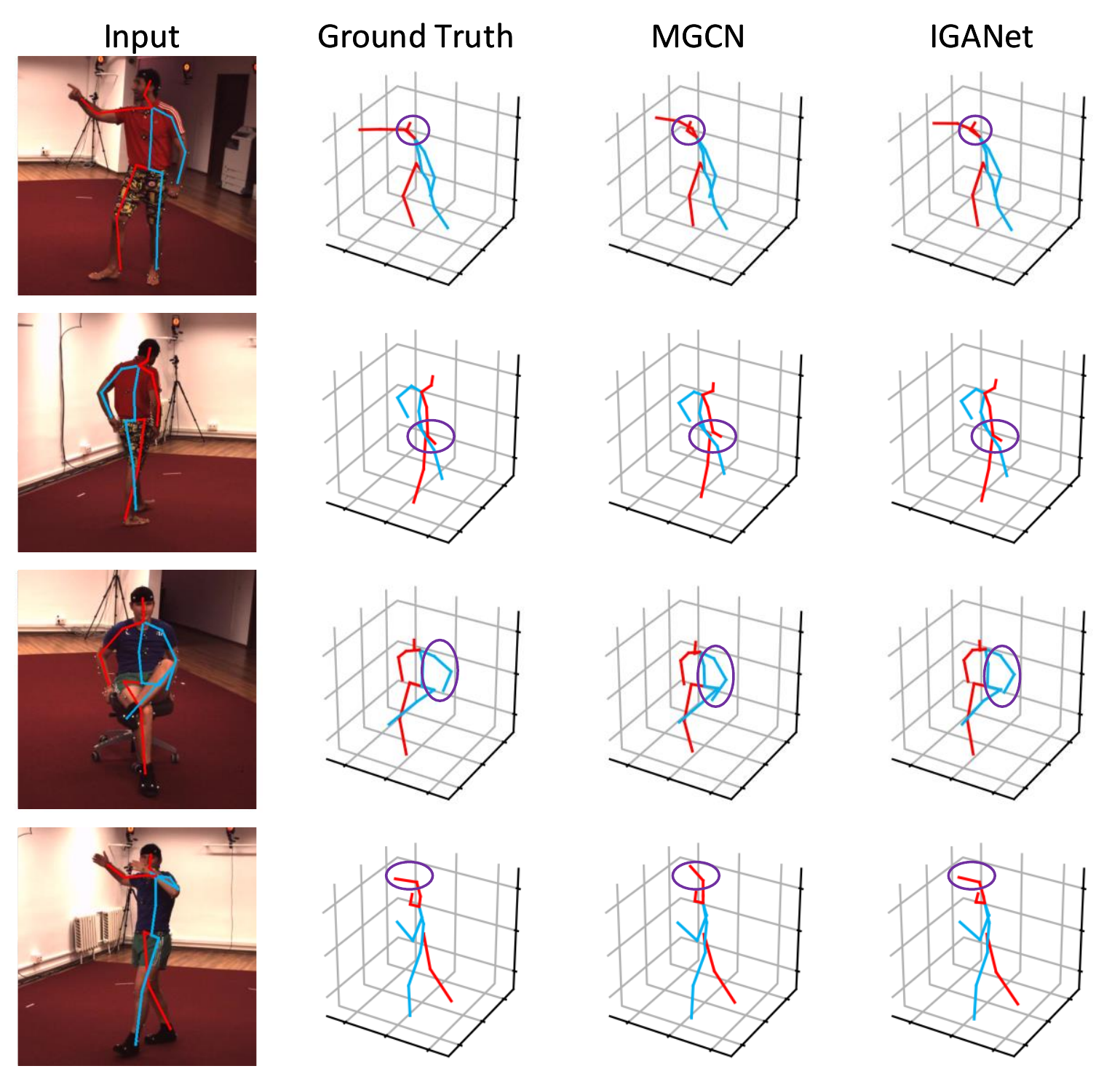}
    \caption{
    Qualitative comparisons with the previous state-of-the-art method (MGCN~\cite{zou2021modulated}) on Human3.6M dataset.}
    \label{fig:vis}
    \vspace{-1mm}
\end{figure}

\noindent \textbf{MPI-INF-3DHP.}
% To examine the generalization ability of IGANet, we directly test our model trained with Human3.6M on the MPI-INF-3DHP dataset.
% We examine IGANet pre-trained on Human3.6M on the test set of MPI-INF-3DHP to examine the generalization ability of IGANet.
We evaluate the generalization ability of IGANet by testing its performance on the test set of MPI-INF-3DHP after pre-training on Human3.6M.
% Comparision with previous stat-of-the-art methods are shown in Table~\ref{tab:3dhp}.
% Our model outperforms most models by a large margin and is comparable to MGCN~\cite{zou2021modulated}.
Comparison results in Table~\ref{tab:3dhp} show our model achieves the best performance over all metrics, which demonstrates the superior generalization ability of our model in unseen scenarios.

\vspace{-1mm}
\subsection{Ablation Study}
% To study the design choices of the IGA module and the advantages of the uMLP module, 
% In order to study the design choices of the IGA module, as well as the advantages of the uMLP module, 
To investigate the design choices of IGANet, 
experiments are conducted on Human3.6M for analysis and verification. 
% We take 2D poses extracted by CPN~\cite{chen2018cascaded} as inputs and use MPJPE as the evaluation metric.
The 2D pose extracted by CPN~\cite{chen2018cascaded} is used as input.
The results are reported in Table~\ref{tab:ab_study}. 

\noindent \textbf{IGA.}
The combination of attention block and original MLP in transformer is adopted as the baseline.
After introducing the GCN block, the performance is improved.
Further, we try to make connections between GCN and attention blocks. 
We find that either injecting the skeleton information from GCN block into attention block (G2A) or introducing global information from attention block to GCN block (A2G) can boost the performance.
% On the one hand, we inject the skeleton information from GCN block into attention block (G2A), the prediction error decreases 0.7mm MPJPE (from 49.4mm to 48.7mm).
% On the other hand, when global information from attention block is introduced to GCN block (A2G), the performance also can be improved (from 49.4mm to 48.8mm).
Finally, the information from GCN and attention blocks are interweaved (both G2A and A2G), and the performance achieves the best (48.3 mm). % MPJPE.
% On this basis, we finally introduce global perception from attention into GCN, and the performance improves to 48.3mm.
% Finally, we introduce global information from attentions into GCNs, and the performance improves to 48.3mm.
% Finally, we both adopt A2G and G2A to achieve the Interweaved graph and attention module, 

\noindent \textbf{uMLP.}
% As shown in Table~\ref{tab:ab_study}, the original MLP often leads to inferior performance, which indicates the effectiveness of our uMLP module.  
% In order to verify the effect of uMLP module, we replaced uMLP with the original MLP layer in the transformer, 
% and found that the performance would drop by 0.15 mm~0.3 mm, which further proves the simplicity and effectiveness of uMLP. 
% In order to verify the effectiveness of the uMLP module, we replace the original MLP layer in the transformer with uMLP.
% Experimental results show that the performance is significantly improved, which further proves the effectiveness of the designed uMLP.
For some model designs in Table~\ref{tab:ab_study}, we replace the original MLP in transformer with our designed uMLP module.
Experimental results show that uMLP module can better learn sleletal representiations, which further proves its effectiveness.   
% Experimental results show that 
% Experimental results show that the original MLP often leads to inferior performance, which indicates the effectiveness of our uMLP module.  

\vspace{-1mm}
\subsection{Qualitative Results}
Fig.~\ref{fig:vis} shows visualization results on Human3.6M dataset.
% Qualitative comparison with MGCN~\cite{zou2021modulated} on Human3.6M test dataset is shown in Fig.~\ref{fig:vis}.
Our IGANet is able to predict 3D poses that are close to ground truth under various actions performed by different people.
% For example, the sitting pose in row three are better predicted by our methods. 
Compared with the MGCN~\cite{zou2021modulated}, IGANet can predict more reasonable and accurate 3D poses.
For example, in the third row, the 3D pose of the sitting action predicted by our method is close to the ground truth. As a comparison, the left arm predicted by MGCN is at a lower position.
% Compared with MGCN\cite{zou2021modulated}, IGANet can predict more reasonable and accurate 3D poses in various poses.
% Compared with the MGCN, IGANet can predict more reasonable and accurate 3D poses.

\vspace{-1mm}
\section{Conclusion}
\vspace{-1mm}

In this paper, we present a novel Interweaved Graph and Attention Network (IGANet) for 3D human pose estimation.
The key idea is to interweave GCNs and attentions to better capture both global and local relations of human body joints. 
To interchange complementary information with each other, we introduce an IGA module with a bidirectional guidance mechanism. 
Moreover, a uMLP module is designed to gather multi-granularity information. 
Experimental results demonstrate the effectiveness of our IGANet for estimating 3D human pose.

{
\small
\bibliographystyle{IEEEbib}
\bibliography{ref}
}

\end{document}